%% file: main.tex
\documentclass[10pt,twocolumn,letterpaper]{article}

\usepackage{iccv}
\usepackage{times}
\usepackage{epsfig}
\usepackage{graphicx}
\usepackage{amsmath}
\usepackage{amssymb}
\usepackage{xcolor}
\usepackage{multirow,tabularx}
\usepackage[percent]{overpic}
\usepackage{gensymb}
\usepackage{pifont}
\usepackage{adjustbox}
\usepackage{array}
\usepackage[accsupp]{axessibility}

\usepackage[pagebackref=true,breaklinks=true,letterpaper=true,colorlinks,bookmarks=false]{hyperref}

\iccvfinalcopy %

\def\iccvPaperID{6149} %

\input{definitions}

\begin{document}
	
	\title{imGHUM: Implicit Generative Models of 3D Human Shape and Articulated Pose}
	
	\author{Thiemo Alldieck\textsuperscript{*} \and Hongyi Xu\textsuperscript{*} \and Cristian Sminchisescu}
	
	\makeatletter
	\let\@oldmaketitle\@maketitle%
	\renewcommand{\@maketitle}{
		\@oldmaketitle%
		\centering
		\vspace{-7mm}
		{\bf Google Research}\\
		{\tt\small \{alldieck,hongyixu,sminchisescu\}@google.com}
		\vspace{6mm}
	}
	\makeatother
	
	\maketitle

	\begin{abstract}
		\vspace{-2mm}
		We present {\bf imGHUM}, the first holistic generative model of 3D human shape and articulated pose, represented as a signed distance function.
		In contrast to prior work, we model the full human body implicitly as a function zero-level-set and without the use of an explicit template mesh.
		We propose a novel network architecture and a learning paradigm, which  make it possible to learn a detailed implicit generative model of human pose, shape, and semantics, on par with state-of-the-art mesh-based models.
		Our model features desired detail for human models, such as articulated pose including hand motion and facial expressions, a broad spectrum of shape variations, and can be queried at arbitrary resolutions and spatial locations.
		Additionally, our model has attached spatial semantics making it straightforward to establish correspondences between different shape instances, thus enabling applications that are difficult to tackle using classical implicit representations.
		In extensive experiments, we demonstrate the model accuracy and its applicability to current research problems.
		\blfootnote{\textsuperscript{*} The first two authors contributed equally.}
		\vspace{-4mm}
	\end{abstract}
	
	\section{Introduction}
	
	\label{sec:introduction}

\input{sections/introduction}
	
	\subsection{Related Work}
	\label{sec:related}
	\input{sections/related}

	\section{Methodology}
	\label{sec:method}

\input{sections/method}

	\section{Experiments}
	\label{sec:experiments}
	\input{sections/experiments}

	\section{Discussion and Conclusion}
	\label{sec:conclusion}
	\input{sections/conclusion}

	{\small
		\bibliographystyle{ieee_fullname}

\input{main.bbl}
	}
	
	\newpage
	
	\section*{Supplementary Material}
	
	\input{sections/suppl}

\end{document}

%% file: definitions.tex
\newcommand{\bb}{\mathbf{b}}
\newcommand{\cc}{\mathbf{c}}
\newcommand{\centers}{\mathbf{j}}

\newcommand{\pp}{\mathbf{p}}
\newcommand{\nn}{\mathbf{n}}
\newcommand{\ee}{\mathbf{e}}
\newcommand{\vv}{\mathbf{v}}
\newcommand{\ww}{\mathbf{w}}
\newcommand{\mm}{\mathbf{m}}

\newcommand{\CC}{\mathbf{C}}

\newcommand{\RR}{\mathbf{R}}
\newcommand{\BB}{\mathbf{B}}
\newcommand{\TT}{\mathbf{T}}
\newcommand{\XX}{\mathbf{X}}
\newcommand{\YY}{\mathbf{Y}}

\newcommand{\balpha}{\boldsymbol{\alpha}}
\newcommand{\bbeta}{\boldsymbol{\beta}}
\newcommand{\btheta}{\boldsymbol{\theta}}

\newcommand{\cmark}{\ding{51}}%
\newcommand{\xmark}{\ding{55}}%
\newcolumntype{R}[2]{%
    >{\adjustbox{angle=#1,lap=\width-(#2)}\bgroup}%
    l%
    <{\egroup}%
}
\newcommand*\rot{\multicolumn{1}{R{35}{1em}}}

\newcommand\blfootnote[1]{%
  \begingroup
  \renewcommand\thefootnote{}\footnote{#1}%
  \addtocounter{footnote}{-1}%
  \endgroup
}

%% file: sections/introduction.tex
Mathematical models of the human body have been proven effective in a broad variety of tasks.
In the last decades models of varying degrees of realism have been successfully deployed e.g.\ for 3D human motion analysis \cite{sminchisescu2002human}, 3D human pose and shape reconstruction \cite{kanazawaHMR18,Zanfir2020Weakly}, personal avatar creation \cite{alldieck2019learning,zhi2020texmesh}, medical diagnosis and treatment \cite{Hesse:PAMI:2019}, or image synthesis and video editing \cite{zanfir2020human,Jain:2010:MovieReshape}.
Modern statistical body models are typically learnt from large collections of 3D scans of real people, which are used to capture the body shape variations among the human population.
Dynamic scans, when available, can be used to further model how different poses affect the deformation of the muscles and the soft-tissue of the human body.

\begin{figure}
    \centering
    \includegraphics[width=1\linewidth]{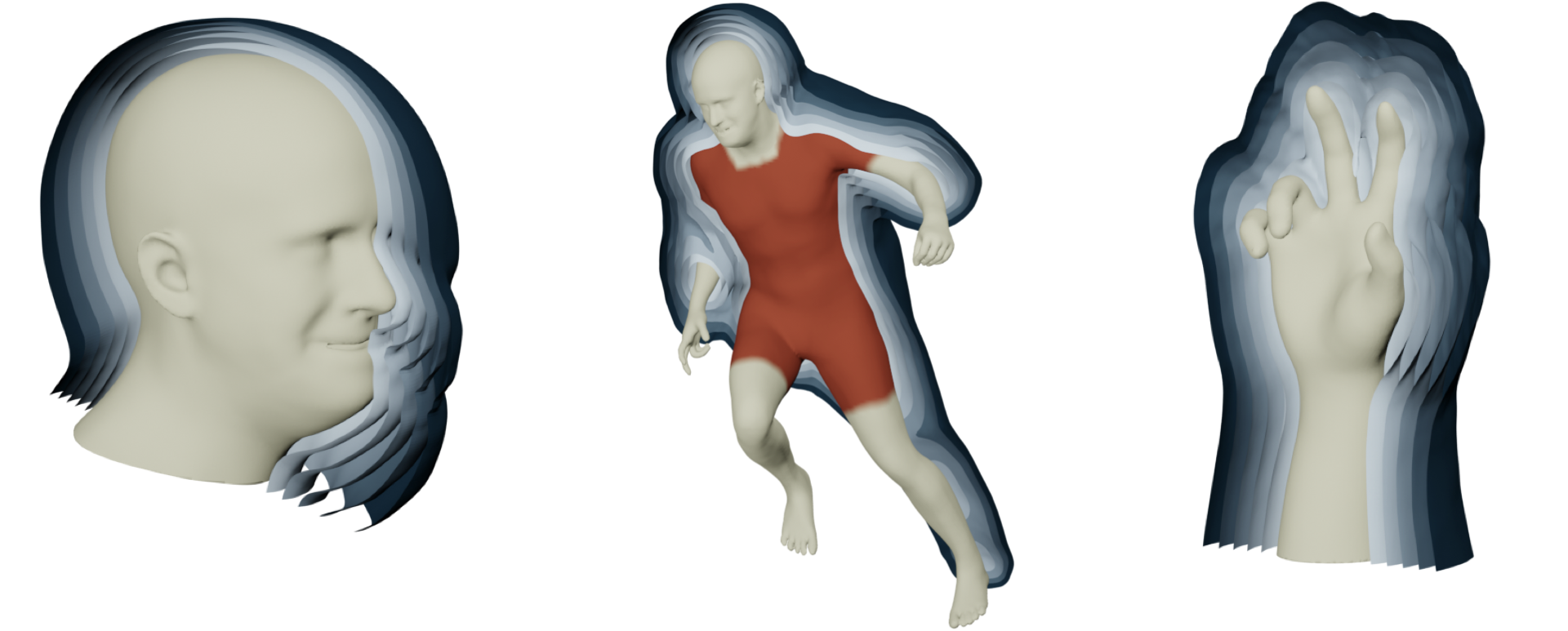}%
    \vspace{-.5mm}
    \caption{{\bf imGHUM} is the first parametric full human body model represented as an implicit signed distance function. {\bf imGHUM} successfully models broad variations in pose, shape, and facial expressions. The level sets of {\bf imGHUM} are shown in blue-scale.}
    \label{fig:teaser}
    \vspace{-4mm}
\end{figure}

The recently released GHUM model~\cite{xu2020ghum} follows this methodology by describing the human body, its shape variation, articulated pose including fingers, and facial expressions as a moderate resolution mesh based on a low-dimensional, partly interpretable parameterization.
In the deep learning literature GHUM and similar models \cite{smpl2015loper,joo2018total} are typically used as fixed function layers.
This means that the model is parameterized with the output of a neural network or some other non-linear function, and the resulting mesh is used to compute the final function value.
While this approach works well for several tasks, including, more recently, 3D reconstruction, the question of how to best represent complex 3D deformable and articulated structures is open.
Recent work dealing with the 3D visual reconstruction of general objects aimed to represent the output not as meshes but as implicit functions \cite{i_OccNet19, i_DeepSDF, i_IMGAN19, i_DeepLevelSet}.
Such approaches thus describe surfaces by the zero-level-set (decision boundary) of a function over points in 3D-space.
This has clear benefits as
the output is neither constrained by a template mesh topology, nor is it discretized and thus of fixed spatial resolution.

In this work, we investigate the possibility to learn a data-driven statistical body model as an implicit function. Given the maturity of state of the art explicit human models, it is crucial that an equivalent implicit representation maintains their key, attractive properties -- representing comparable variation in shape and pose and similar level of detail.
This is challenging since recently-proposed implicit function networks tend to produce overly smooth shapes and fail for articulated humans \cite{chibane20ifnet}.
We propose a novel network architecture and a learning paradigm that enable, for the first time, constructing detailed generative models of human pose, shape, and semantics, represented as Signed Distance Functions (SDFs) (see fig.~\ref{fig:teaser}).
Our multi-part architecture focuses on difficult to model body components like hands and faces.
Moreover, imGHUM models its neighborhood through distance values, enabling e.g.\ collision tests.
Our model is not bound to a specific resolution and thus can be easily queried at arbitrary locations.
Being template-free further paves the way to our ultimate goal to fairly represent diversity of mankind, including disabilities which may not be always well covered by a generic template of standard topology.
Finally, in contrast to recent implicit function networks, our model additionally carries on the explicit semantics of mesh-based models.
Specifically, our implicit function also returns correspondences to a canonical representation near and on its zero-level-set, enabling e.g.\ texturing or body part labeling.
This holistic approach is novel and significantly more difficult to produce, as can be noted in prior work which could only demonstrate individual properties, \cf tab.\ \ref{tab:features}. Our contribution -- and the key to success -- stems from the novel combination of adequate, generative latent representations, network architectures with fine grained encoding, implicit losses with attached semantics, and the consistent aggregation of multi-part components.
Besides extensive evaluation of 3D deformable and articulated modeling capabilities, we also demonstrate surface completion using imGHUM and give an outlook to modeling varying topologies.
Our models are available for research~\cite{imghumcode}.

\begin{table}
    \begin{center}
    \scriptsize
    \begin{tabular}{cccccccc|l} \rot{generative pose} & \rot{generative shape}  & \rot{generative hands} & \rot{gen. expression} & \rot{interpolation} & \rot{signed distances} & \rot{semantics} & \rot{continuous rep.} & \\
    \hline\hline
    \cmark   & \cmark & \cmark & \cmark & \cmark & \xmark & \cmark &  \xmark  & GHUM \cite{xu2020ghum} \\
    \hline
    \xmark & \xmark  & \xmark & \xmark & \xmark & \xmark & \xmark & \cmark  & IF-Net \cite{chibane20ifnet} \\
    \xmark & \xmark & \xmark & \xmark & \cmark & \cmark & \xmark & \cmark  & IGR \cite{gropp2020igr} \\
    \cmark   & \xmark & \xmark & \xmark & \cmark & \xmark & \xmark & \cmark  & NASA \cite{deng2019nasa} \\
    \hline
     \cmark   & \cmark & \cmark & \cmark & \cmark & \cmark & \cmark & \cmark   & \textbf{imGHUM}
    \end{tabular}
    \end{center}
    \vspace{-1.5mm}
    \caption{Comparison of different approaches to model human bodies. GHUM is meshed-based and thus discretized. IGR only allows for shape interpolation. NASA lacks generative capabilities for shape, hands, and facial expressions and only returns occupancy values. Only imGHUM combines all favorable properties.}
    \label{tab:features}
    \vspace{-3mm}
\end{table}

%% file: sections/related.tex
We review developments in 3D human body modeling, variants of implicit function networks, and applications of implicit function networks for 3D human reconstruction.\\

\vspace{-3mm}

\noindent{\bf Human Body Models.}
Parametric human body models based on geometric primitives have been proposed early on \cite{thalmann1996fast}
and successfully applied e.g.\ for human reconstruction from video data \cite{plankers2001articulated,sminchisescu2002human, sminchisescu2006learning}. 
SCAPE \cite{fscape} was one of the first realistic large scale data-driven human body models. Later variants inspired by blend skinning \cite{hirshberg2012blendscape} modeled correlations between body shape and pose \cite{hasler2009statistical}, as well as soft-tissue dynamics \cite{pons2015dyna}.
SMPL variants \cite{smpl2015loper,joo2018total,SMPL-X:2019,STAR:2020} are also popular parametric body models, with linear shape spaces, compatible with standard graphics pipelines and offering good full-body representation functionality.
GHUM is a recent parametric model \cite{xu2020ghum} that represents the full body model using deep non-linear models -- VAEs for shape and normalizing flows for pose, respectively -- with various trainable parameters, learned end-to-end.  
In this work, we rely on GHUM to build our novel implicit model.
Specifically, besides the static and dynamic 3D human scans in our dataset, we also rely on GHUM (1) to represent the latent pose and shape state of our implicit model, (2) to generate supervised training data in the form of latent pose and shape codes with associated 3D point clouds, sampled from the underlying, posed, GHUM mesh. \\

\vspace{-3mm}

\noindent{\bf Implicit Function Networks (IFNs)} have been proposed recently \cite{i_OccNet19, i_DeepSDF, i_IMGAN19, i_DeepLevelSet}.
Instead of representing shapes as meshes, voxels, or point clouds, IFNs learn a shape space as a function of a low-dimensional global shape code and a 3D point.
The function either classifies the point as inside/outside \cite{i_OccNet19, i_IMGAN19} (occupancy networks), or returns its distance to the closest surface \cite{i_DeepSDF} (distance functions).
The global shape is then defined by the decision boundary or the zero-level-set of this function.

Despite advantages over mesh- and voxel-based representations in tasks like e.g.\ 3D shape reconstruction from partial views or given incomplete data, initial work has limitations. First, while the models can reliably encode rigid axis-aligned shape prototypes, they often fail for more complex shapes.
Second, the reconstructions are often overly smooth, hence they lack detail.
Different approaches have been presented to address these.
Part-based models \cite{i_StructuredImplicitFunctions, jiang2020local, genova2020local} assemble a global shape from smaller local models.
Some methods do not rely on a global shape code but on features computed from convolving with an input observation \cite{chibane20ifnet, chibane2020ifnet_texture, Peng2020convocc, chibane2020ndf}.
Others address such limitations by changing the learning methodology: tailored network initialization \cite{atzmon2020sal} and point sampling strategies \cite{xu2020ladybird}, or second-order losses \cite{gropp2020igr,sitzmann2020siren} have been proposed towards this end.
We found the latter to be extremely useful and rely on similar losses in this work.\\

\vspace{-3mm}
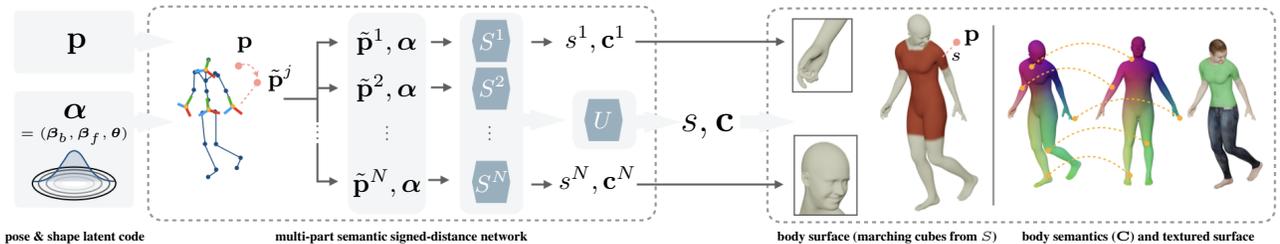
\begin{figure*}
    \centering
    \begin{overpic}[width=0.98\linewidth]{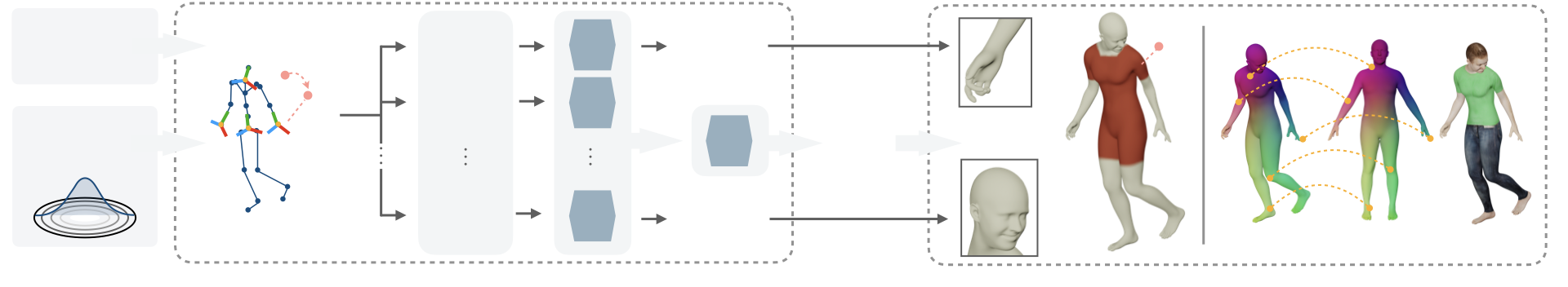} %
        \put(4.5,9.5){\large$\balpha$}
        \put(1.2,8){\tiny$=(\bbeta_b, \bbeta_f, \btheta)$}
        \put(4.8,15){\large$\pp$}
        \put(18,15){\small$\pp$}
        \put(20.4,12){\small$\tilde\pp^j$}
        \put(52.5,8.5){\Large$s,\cc$}
        
        \put(27.3, 14.8){$\tilde\pp^1,\balpha$}
        \put(36.8, 14.8){\color{white}\small$S^1$}
        \put(27.3,11.2){$\tilde\pp^2,\balpha$}
        \put(36.8,11.2){\color{white}\small$S^2$}
        
        \put(27,3.8){$\tilde\pp^N,\balpha$}
        \put(36.5,3.8){\color{white}\small$S^N$}
        
        \put(45.7,8.7){\color{white}\small$U$}
        
        \put(43.5,15){$s^1,\cc^1$}
        \put(43,4){$s^N,\cc^N$}
        
        \put(74.5,15.5){\small$\pp$}
        \put(73.5,14){\scriptsize $s$}
        
        \put(0,0){\sffamily \tiny \bf pose \& shape latent code}
        \put(21,0){\sffamily \tiny \bf multi-part semantic signed-distance network}
        \put(60,0){\sffamily \tiny \bf body surface (marching cubes from $S$)}
        \put(79,0){\sffamily \tiny \bf body semantics ($\CC$) and textured surface}
    \end{overpic}%
    \vspace{1mm}
    \caption{Overview of imGHUM. We compute the signed distance $s=S(\pp,\balpha)$ and the semantics $\cc=\CC(\pp,\balpha)$ of a spatial point $\pp$ to the surface of an articulated human shape defined by the generative latent code $\balpha$. Using an explicit skeleton,
    we transform the point $\pp$ into the normalized coordinate frames as $\{\tilde\pp^j\}$ for $N=4$ sub-part networks, modeling body, hands, and head.
    Each sub-model $\{S^j\}$ represents a semantic signed-distance function. The sub-models are finally combined consistently using an MLP $U$ to compute the outputs $s$ and $\cc$ for the full body. %
    Our multi-part pipeline builds a full body model as well as sub-part models for head and hands, jointly, in a consistent training loop. 
    On the right, we visualize the zero-level-set body surface extracted with marching cubes and the implicit correspondences to a canonical instance given by the output semantics. The semantics allows e.g.\ for surface coloring or texturing.}
    \label{fig:method_overwiew}
    \vspace{-3mm}
\end{figure*}
\noindent{\bf IFNs for Human Reconstruction.}
Recently implicit functions have been explored to reconstruct humans.
Huang et al.\ \cite{huang2018deep} learn an occupancy network that conditions on image features in a multi-view camera setup.
Saito et al.~\cite{saito2019pifu} use features from a single image and an estimated normal image \cite{saito2020pifuhd} together with depth values along camera rays as conditioning variables.
ARCH \cite{huang2020arch} combines implicit function reconstruction and explicit mesh-based human models to represent dressed people.
Karunratanakul et al.\ \cite{karunratanakul2020grasping} propose to use SDFs to learn human grasps and augment their SDFs output with sparse regional labels.
Similarly to us, Deng et al.\ \cite{deng2019nasa} represent a pose-able human subject as a number of binary occupancy functions modeled in a kinematic structure.
In contrast to our work, this framework is restricted to a single person and the body is only coarsely approximated, lacking facial features and hand detail.
Also related, SCANimate \cite{Saito:CVPR:2021} builds personalized avatars from multiple scans of a single person.
Concurrent to our work, LEAP \cite{LEAP:CVPR:21} learns an occupancy model of human shape and pose also without hand poses, expressions, or semantics.
In this work we aim for a full implicit body model, featuring a large range of body shapes corresponding to diverse humans and poses, with detailed hands, and facial expressions.

%% file: sections/method.tex
In this section, we describe our models and the losses used for training.
We introduce two variants: a single-part model that encodes the whole human in a single network and a multi-part model.
The latter constructs the full body from the output superposition of four body part networks.\\

\vspace{-3mm}

\noindent{\bf Background.} We rely on neural networks and implicit functions to generate 3D human shapes and articulated poses. 
Given a latent representation $\balpha$ of the human shape and pose, together with an underlying probability distribution, we model the posed body
as the zero iso-surface decision boundaries of Signed Distance Functions (SDFs) given by deep feed-forward neural networks.
A signed distance $S(\pp, \balpha)\in \RR$ is a continuous function which, given an arbitrary  
spatial point $\pp \in \RR^3$, outputs the shortest distance to the surface defined by $\balpha$, where 
the sign indicates the inside (negative) or outside (positive) side w.r.t.\ the surface. 
The posed human body surface is implicitly given by $S(\cdot, \balpha) = 0$.

GHUM~\cite{xu2020ghum} represents the human model as an articulated mesh $\XX (\balpha)$. GHUM has a minimally-parameterized skeleton with $J=63$ joints ($124$ Euler angle DOFs), and 
skinning deformations, explicitly sensitive to the pose kinematics $\btheta \in \RR^{124}$.
A kinematic prior based on normalizing flows defines the distribution of valid poses \cite{Zanfir2020Weakly}.
Each kinematic pose $\btheta$ represents a set of joint transformations $\TT(\btheta, \centers)\in \RR^{J\times 3\times 4}$ from 
the neutral to a posed state, where $\centers\in \RR^{J\times 3}$ are the joint centers that are dependent on the neutral body shape.
The statistical body shapes are modeled using a nonlinear embedding $\bbeta_b \in \RR^{16}$.
In addition to skeleton articulation, a nonlinear latent code $\bbeta_f\in\RR^{20}$ drives facial expressions. The implicit model we design here shares the same probabilistic latent representation as GHUM, $\balpha=(\bbeta_b, \bbeta_f, \btheta)$, but in contrast to computing an articulated mesh, we estimate a signed distance value $s=S(\pp, \balpha)$ for each arbitrary spatial point $\pp.$

\subsection{Models and Training}
Given a collection of full-body human meshes $\YY$, together with the corresponding GHUM encodings $\balpha=(\bbeta_b, \bbeta_f, \btheta)$, our goal is to learn a MLP-based SDF representation $S(\pp,\balpha)$ so that it approximates the shortest signed distance to $\YY$ for any query point $\pp$. Note that $\YY$ could be arbitrary meshes, such as raw human scans, mesh registrations, or samples drawn from the GHUM latent space. The zero iso-surface $S(\cdot, \balpha) = 0$ is sought to preserve all geometric detail in $\YY$, including body shapes and poses, hand articulation, and facial expressions.\\ 

\vspace{-3mm}

\noindent{\bf Single-part Network.}
\label{sec:singlepart} We formulate one global neural network that decodes $S(\pp, \balpha)$ for a given latent code $\balpha$ and a spatial point $\pp$.  Instead of pre-computing the continuous SDFs from point samples as in DeepSDF~\cite{i_DeepSDF}, we 
train a MLP network $S(\pp, \balpha; \omega)$ with weights $\omega$, similar in spirit to IGR~\cite{gropp2020igr}, to output a solution to the Eikonal equation 
\begin{equation}
\label{eq:eikonal}
    \small
    \|\nabla_\pp S(\pp, \balpha; \omega) \| = 1,
\end{equation}
where $S$ is a signed distance function that vanishes at the surface $\YY$ with gradients equal to surface normals. 
Mathematically, we formulate our total loss as a weighted combination of
\vspace{-1mm}
\begin{gather}
    \footnotesize
    \label{eq:onloss}
    L_{o}(\omega) = \frac{1}{|O|}\sum_{i\in O}(|S(\pp_i, \balpha)| + \| \nabla_{\pp_i}S(\pp_i, \balpha) - \nn_i \|) \\
    \footnotesize
    \label{eq:offnormloss}
    L_{e}(\omega) = \frac{1}{|F|}\sum_{i\in F}(\|\nabla_{\pp_i}S(\pp_i, \balpha)\| - 1)^2 \\
    \footnotesize
    \label{eq:labelloss}
    L_{l}(\omega) = \frac{1}{|F|}\sum_{i\in F}\text{BCE}(l_i, \phi(k S(\pp_i, \balpha))),
\end{gather}
where $\phi$ is the sigmoid function,
$O$ are surface samples from $\YY$ with normals $\nn$, and $F$ are off surface samples with inside/outside labels $l$, consisting of both uniformly sampled points within a bounding box and sampled points near the surface. 
The first term $L_{o}$ encourages the surface samples to be on the zero-level-set and the SDF gradient to be equal to the given surface normals $\nn_i$. The Eikonal loss $L_{e}$ is derived from \eqref{eq:eikonal} where the SDF is differentiable everywhere with gradient norm $1$. 
We obtain the SDF gradient $\nabla_{\pp_i}S(\pp_i, \balpha)$ analytically via network back-propagation. In practice, we also find it useful to include a binary cross-entropy error (BCE) loss $L_{l}$ for off-the-surface samples, where $k$ controls the sharpness of the decision boundary.
We use $k=10$ in our experiments.
Our training losses only require surface samples with normals and inside/outside labels for off-surface samples.
Those are much easier and faster to obtain than pre-computing ground truth SDF values.

Recent work suggests that standard coordinate-based MLP networks encounter difficulties in learning high-frequency functions, a phenomenon referred to as \emph{spectral bias}~\cite{rahaman2019spectral, tancik2020fourfeat}.
To address this limitation, inspired by~\cite{tancik2020fourfeat}, we therefore encode our samples using the basic Fourier mapping $\ee_i = [\sin(2\pi\tilde\pp_i), \cos(2\pi\tilde\pp_i)]^\top,$ where we first unpose the samples with the root rigid transformation $\TT_{0}^{-1}$ and normalize them into $[0,1]^3$ with a shared bounding box $\BB = [\bb_{min}, \bb_{max}]$, as
\begin{gather}
    \small
    \tilde\pp_i = \frac{\TT_{0}^{-1}(\btheta, \centers)[\pp_i, 1]^\top - \bb_{min}}{\bb_{max}- \bb_{min}}.  
\end{gather}
Note that our SDF is defined w.r.t.\ the original meshes $\YY$ and therefore we do not unpose and scale the sample normals.
Also, the loss gradients are derived w.r.t.\ $\pp_i$.\\

\vspace{-3mm}

\noindent{\bf Multi-part Network.}
\label{sec:multipart}
Our single-part network represents well the global geometric features for various human body shapes and kinematic poses.
However, despite its spatial encoding, the network still has difficulties capturing facial expressions and articulated hand poses, where the SDF has local high-frequency variations. 
To augment geometric detail on face and hands regions, we therefore propose a multi-part network 
that decomposes the human body into $N=4$ local regions, i.e. the head, left and right hand, and the remaining body, respectively.
This significantly reduces spectral frequency variations within each local region allowing the specialized single-part networks to capture local geometric detail.
A consistent full-body SDF $S(\pp,\balpha)$ is composed from the local single-part SDF network outputs $s^j = S^j(\pp, \balpha), j \in \{ 1, \ldots, N\}$.

We follow the training protocol described in \S\ref{sec:singlepart} for each local sub-part network with surface and off-surface samples within a bounding box $\BB^j$ defined for each part.
Note that we use the neck and wrist joints as the the root transformation for the head and hands respectively.
In GHUM, the joint centers $\centers$ are obtained as a function given the neutral body shapes $\bar{\XX}(\bbeta_b)$.
However, $\bar{\XX}$ is not explicitly presented in our implicit representation. 
Therefore, we build a nonlinear joint regressor from $\bbeta_b$ to $\centers$, which is trained, supervised, using GHUM's latent space sampling.   

In order to fuse the local SDFs into a consistent full-body SDF, while at the same time preserving local detail, we merge the last hidden layers of the local networks using an additional light-weight MLP $U$.
To train the combined network, a sample point $\pp_i$, defined for the full body, is transformed into the $N$ local coordinate frames using $\TT^j_{0}$ and then passed to the single-part local networks, see fig.~\ref{fig:method_overwiew}.
The union SDF MLP then aggregates the shortest distance to the full body among the local distances.
We apply our losses to the union full-body SDF as well, to ensure that the output for full body satisfies the SDF property \eqref{eq:eikonal}. Our multi-part pipeline produces sub-part models and a full-body one, trained jointly and leveraging data correlations among different body components. 

Our spatial point encoding $\ee_i$ requires all samples $\pp$ to be inside the bounding box $\BB$, which otherwise might result in periodic SDFs due to sinusoidal encoding.
However, a point sampled from the full body is likely to be outside of a sub-part's local bounding box $\BB^j$. Instead of clipping or projecting to the bounding box, we augment our encoding of sample $\pp_i$ for sub-part networks $S^j$ as $\ee^j_i = [\sin(2\pi\tilde\pp_i^j), \cos(2\pi\tilde\pp_i^j), \tanh(\pi(\tilde\pp_i^j - 0.5))]^\top$, where the last value indicates the relative spatial location of the sample w.r.t.\ the bounding box.
If a point $\pp_i$ is outside the bounding box $\BB^j$, the union SDF MLP will learn to ignore $S^j(\pp^j_i, \balpha)$ for the final union output.  \\ 

\vspace{-3mm}

\noindent{\bf Implicit Semantics.}\label{sec:sematics} In contrast to explicit models like GHUM, implicit functions do not naturally come with point correspondences between different shape instances.
However, many applications, such as pose tracking, texture mapping, semantic segmentation, surface landmarks, or clothing modeling, largely benefit from such correspondences. %
Given an arbitrary spatial point, on or near the surface $\YY$, i.e, $|S(\pp_i, \balpha)| < \sigma$, we are therefore interested to interpret its \emph{semantics}. %
We define the semantics as a 3D implicit function $\CC(\pp, \balpha)  \in \RR^{3}$.
Given a query point $\pp_i$, it returns a correspondence point on a canonical GHUM mesh $\XX(\balpha_0)$ as
\vspace{-3mm}
\begin{equation}
\small
\label{eq:sem}
    \CC(\pp_i, \balpha) = \ww_{i} \vv_{f} (\balpha_0) = \cc_i, \quad\, \pp_i^* = \ww_{i} \vv_{f} (\balpha)
\end{equation}
where $\pp_i^*$ is the closest point of $\pp_i$ in the GHUM mesh $\XX(\balpha)$ 
with $f$ the nearest face and $\ww$ the barycentric weights of the vertex coordinates $\vv_{f}.$
In contrast to alternative semantic encodings, such as 2D texture coordinates, our semantic function $\CC(\pp, \balpha)$ is smooth in the spatial domain without distortion and boundary discontinuities, which favors the learning process, \cf.\ \cite{bhatnagar2020loopreg}.

By definition, implicit SDFs return the shortest distance to the underlying implicit surface for a spatial point whereas implicit semantics associate the query point to its closest surface neighbor. Hence, we consider implicit semantics as highly correlated to SDF learning.
We co-train both tasks with our augmented multi-part network (\S\ref{sec:multipart}) computing both $S(\pp, \balpha)$ and $\CC(\pp, \balpha)$. 
Semantics are trained fully supervised, using an $L_1$ loss for a collection of training sample points near and on the surface $\YY$. 
Due to the correlation between tasks, our network is able to predict both signed distance and semantics, without expanding its capacity.

Using trained implicit semantics, we can e.g.\ apply textures to arbitrary iso-surfaces at level set $|z|\leq \sigma$, reconstructed from our implicit SDF. 
During inference, an iso-surface mesh $S(\cdot, \balpha) = z$ can be extracted using Marching Cubes~\cite{lewiner2003efficient}. 
Then for every generated vertex $\tilde\vv_i$ we query its semantics $\CC(\tilde\vv_i, \balpha).$ 
The queried correspondence point $\CC(\tilde\vv_i, \balpha)$ might not be exactly on the canonical surface and therefore we project it onto $\XX(\balpha_0)$.
Now, we can interpolate the UV texture coordinates and assign them to $\tilde\vv_i.$ 
Similarly, we can also assign segmentation labels or define on-\ or near-surface landmarks.
In fig.~\ref{fig:method_overwiew} (right) we show an imGHUM reconstruction textured and with a binary `clothing' segmentation.
We use the latter throughout the paper demonstrating that our semantics allow the transfer of segmentation labels to different iso-surface reconstructions. Please refer to \S\ref{sec:applications} for more applications of our implicit semantics e.g.\ landmarks or clothed human reconstruction. \\

\vspace{-3mm}
\noindent{\bf Architecture.}
For the single-part network we use a similar feed-forward architecture as DeepSDF~\cite{i_DeepSDF} or IGR \cite{gropp2020igr} with eight $512$-dimensional fully-connected layers. 
To enable higher-order derivatives, we use Swish nonlinear activation \cite{ramachandran2017searching} instead of ReLU. IGR originally proposed SoftPlus, however, we found Swish superior (see tab.~\ref{tab:compare_baselines}).
The multi-part network is composed out of one $8$-layer $256$-dimensional MLP for the body and three $4$-layer $256$-dimensional MLPs for hands and head.
Each sub-network has a skip connection to the middle layer.
The last hidden layers of sub-networks are aggregated in a 128-dimensional fully-connected layer with Swish nonlinear activation, before the final network output. %
The final model features $2.49$ million parameters and performs $4.99$ million FLOPs per point query.\\

\vspace{-3mm}

\noindent{\bf Dataset.}
Our training data consists of a collection of full-body human meshes $\YY$ together with the corresponding GHUM latent code $\balpha,$ where $\XX(\balpha)$ best approximates $\YY.$  For each mesh, we perform Poisson disk sampling on the surface and obtain $|O| = 32$K surface samples, together with their surface normals. In addition, within a predefined $2.2 \times 2.8 \times 2.2 \,\textrm{m}^3$ bounding box centered at the origin, we sample $|F| / 2 = 16$K points uniformly. Another $16$K samples are generated by randomly displacing surface sample points with isotropic normal noise with $\sigma=0.05$m. All off-surface samples are associated with inside/outside labels, computed by casting randomized rays and checking parity. We also label semantics for on and near surface samples, which are drawn with random face indices and barycentric weights of the GHUM mesh and randomly displaced for near-surface samples. With the corresponding face and barycentric weights, semantic labels are generated using \eqref{eq:sem} in a light-weight computation with no need for projection or nearest neighbor search. Each mesh $Y$ is then decomposed into $N=4$ parts and we generate the same number of training samples per body part (we use $\sigma=0.02$m for surface samples near the hands). %

We use two types of human meshes for our imGHUM training. We first randomly sample $75$K poses from H36M and the CMU mocap dataset, with Gaussian sampled body shapes, expressions and hand poses from the GHUM latent priors, where $\YY$ are the posed GHUM meshes. In addition, we collect $35$K human scans, on which we perform As-Conformal-As-Possible (ACAP) registrations~\cite{Yoshiyasu2014} with the GHUM topology and fit GHUM parameters as well. Our human scans include the CAESAR dataset, full body pose scans, as well as close-up head and hand scans. Due to the noise and incompleteness in some of the raw scans we use the registrations for training. We fine-tune imGHUM -- initially trained on GHUM sampling -- using the registration dataset. In this way, imGHUM can capture geometric detail not well represented by GHUM (see tab.~\ref{tab:compghum}).

%% file: sections/experiments.tex
We evaluate imGHUM qualitatively and quantitatively in multiple experiments.
First, we compare imGHUM with its explicit counterpart GHUM (\S\ref{sec:compare_ghum}).
Then, we perform an extensive baseline and ablation study, demonstrating the effect of imGHUM's architecture and training scheme (\S\ref{sec:compare_baselines}). We also build a model to compare to the recent single-subject occupancy model NASA.
Finally, we show the performance of imGHUM on three representative applications demonstrating its usefulness and versatility (\S\ref{sec:applications}).

We report three different metrics. Bi-directional Chamfer-$L_2$ distance measures the accuracy and completeness of the surface (lower is better). Normal Consistency (NC) evaluates estimated surface normals (higher is better). Volumetric Intersection over Union (IoU) compares the reconstructed volume with the ground truth shape (higher is better).
The latter can only be reported for watertight shapes.
Please note that metrics not always correlate with the perceived quality of the reconstructions.
We therefore additionally include qualitative side-by-side comparisons.

\begin{figure*}
    \centering
    \includegraphics[width=1\linewidth]{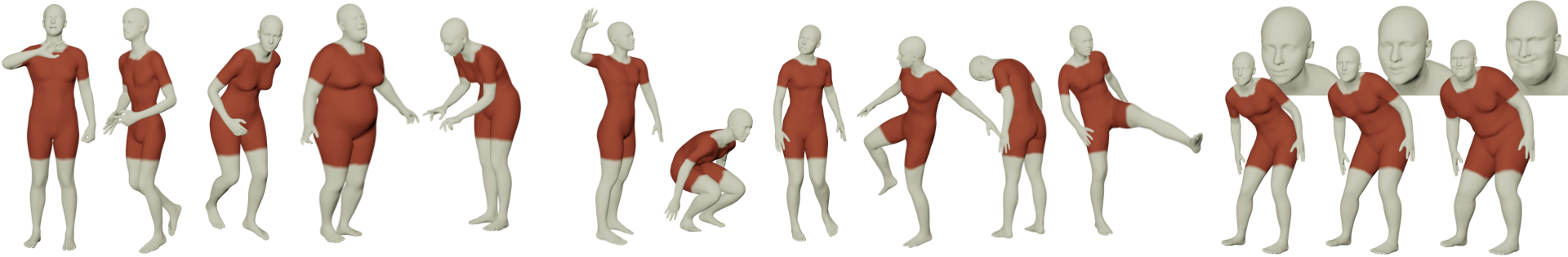}%
    \vspace{-0.5mm}
    \caption{Bodies generated and reconstructed using imGHUM. \emph{Left}: imGHUM with Gaussian sampling of the shape, expression and pose latent space. \emph{Middle}: Reconstructed motion sequence from the CMU mocap dataset~\cite{CMUMotionCap2000} (fixed body shape). \emph{Right}: Body shape and facial expressions latent code interpolation (fixed pose). See supplementary material for more examples.}
    \label{fig:examples}
    \vspace{-3mm}
\end{figure*}

For visualization and numerical evaluation we extract meshes from imGHUM using Marching Cubes~\cite{lewiner2003efficient}.
To this end, we approximate the bounding box of the surface though probing and then run Marching Cubes with a resolution of $256^3$ within the bounding box.
Hereby, the signed distances support acceleration using Octree sampling: we use the highest grid density only near the surface and sample far less frequently away from it.
However, we note that for most applications, such as human reconstruction and collision detection, Marching Cubes are not needed, except only once for the final mesh visualization.

\subsection{Representational Power}
\label{sec:compare_ghum}
In fig.~\ref{fig:examples}, we show reconstructions of a motion capture sequence applied to imGHUM. 
Our model captures well the articulated full-body motion, with consistent body shape for various poses.
By sharing the latent priors with GHUM, imGHUM supports realistic body shape and pose generation (fig.~\ref{fig:examples}, left) as well as smooth interpolation within the shape and expression latent spaces (fig.~\ref{fig:examples}, right).  Our model generalizes well to novel body shapes, expressions, and poses, and has interpretable and decoupled latent representations. 

In tab.~\ref{tab:compghum}, we compare the representation power of imGHUM with the explicit GHUM on our registration test-set. 
imGHUM better captures present detail as numerically demonstrated.
An imGHUM model trained only using GHUM samples captures the body deformation due to articulation less well, indicating that GHUM is a useful surrogate to `synthetically' bootstrap the training of the implicit network, but that real data is important as well.

\begin{figure}
    \centering
    \includegraphics[width=0.9\linewidth]{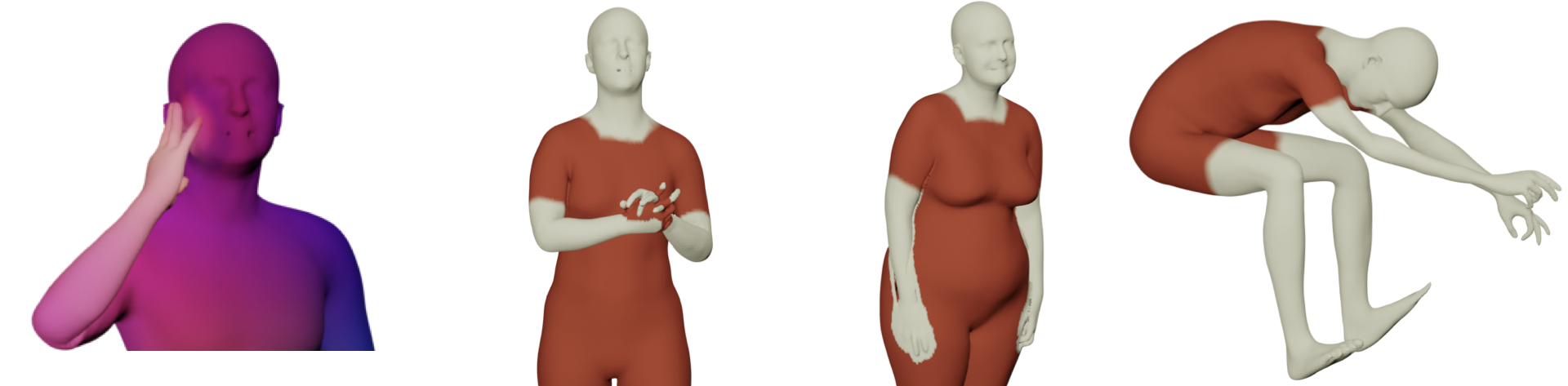}
    \caption{Failure modes. Interpenetration can lead to unwanted shapes and semantics (leaked hand semantics to the cheek). Extreme poses may produce deformed body parts (thin arms).}
    \label{fig:limitation}
    \vspace{-3mm}
\end{figure}

\noindent{\bf Limitations} of imGHUM are sometimes apparent for very extreme pose configurations that have not been covered in the training set, such as anthropometrically invalid poses that are impossible for a human, e.g.\ resulting in self-intersection or by bending joints beyond their anatomical range of motion.
imGHUM produces plausible results for inputs not too far from expected configurations, but the results occasionally feature some defects e.g.\ distorted or incomplete geometry or inaccurate semantics, see fig.~\ref{fig:limitation} for examples.

\subsection{Baseline Experiments}
\label{sec:compare_baselines}

\begin{table}
    \vspace{-1mm}
    \begin{center}
    \footnotesize
    \begin{tabular}{l|ccc}  

    Model & IoU $\uparrow$ & Chamfer $\times 10^{-3}$ $\downarrow$ & NC $\uparrow$\\
    \hline
    \hline
    imGHUM $\ddagger$ &0.900 & 0.071  & 0.977 \\
    GHUM & 0.913 & 0.055  & 0.983 \\
    \hline
    \textbf{imGHUM} & \textbf{0.932} & \textbf{0.040}  & \textbf{0.984}
    \end{tabular}
    \end{center}
    \vspace{-1mm}
    \caption{GHUM comparisons on registration dataset. imGHUM marked with $\ddagger$ is trained only based on GHUM sampling data. }
    \label{tab:compghum}
    \begin{center}
    \footnotesize
    \begin{tabular}{l|ccc}  

    Model & IoU $\uparrow$ & Chamfer $\times 10^{-3}$ $\downarrow$ & NC $\uparrow$\\
    \hline
    \hline
    Autoencoder                     & 0.831 & 2.457 & 0.923 \\ 
    Single-part $\dagger$           & 0.957 & 0.085 & 0.983 \\ 
    Single-part $\oplus$            & 0.958 & 0.070 & 0.983 \\
    Single-part                     & 0.965 & 0.052 & 0.986 \\ 
    Single-part deeper $\dagger$    & 0.961 & 0.070 & 0.984 \\ 
    Single-part deeper              & 0.967 & 0.058 & 0.986 \\ 
    imGHUM $\dagger$                & 0.955 & 0.095 & 0.984 \\
    imGHUM w/o $L_{l}$~\eqref{eq:labelloss}               & 0.966 & 0.051 & 0.988 \\ 
    \hline
    \textbf{imGHUM} & \textbf{0.969} & \textbf{0.036} & \textbf{0.989}
    \end{tabular}
    \end{center}
    \vspace{-1mm}
    \caption{Numerical comparison with baselines. Models marked with $\dagger$ don't use the Fourier input mapping. $\oplus$ marks Softplus activation as in \cite{gropp2020igr}.}
    \label{tab:compare_baselines}

    \begin{center}
    \footnotesize
    \begin{tabular}{l|ccc}  
    \multirow{2}{*}{Model} & IoU $\uparrow$ & Ch. $\times 10^{-3}$ $\downarrow$ & NC $\uparrow$ \\
     & {\scriptsize Head / Hands} & {\scriptsize Head / Hands} & {\scriptsize Head / Hands} \\
    \hline
    \hline
    Single-part             & 0.967 / 0.818 & 0.010 / 0.201 & 0.937 / 0.790 \\ 
    Single-part deep.      & 0.968 / 0.832 & 0.011 / 0.271 & 0.938 / 0.811 \\ 
    \hline
    \textbf{imGHUM} & \textbf{0.976} / \textbf{0.929} & \textbf{0.007} / \textbf{0.031} & \textbf{0.944} / \textbf{0.934}
    \end{tabular}
    \end{center}
    \vspace{-1mm}
    \caption{Unidirectional metrics (GT to generated mesh) for critical body parts. Our multi-part architecture significantly improves the head and hand reconstruction accuracy.}
    \label{tab:compare_baselines_parts}
    \vspace{-4mm}
\end{table}

In the next section, we compare imGHUM to various baselines inspired by recent work.
The first is an auto-encoder, where the encoder side is PointNet++ \cite{qi2017pointnet++} and the decoder is our single-part network.
The idea is to let the network find the best representation instead of pre-computing a low dimensional representation. 
In practice this means that latent codes are not interpretable.
Further, we experiment with our single part network without Fourier input mapping, largely following the training scheme proposed by IGR \cite{gropp2020igr}. We also use input mapping and finally trained a deeper single-part network variant (10 layers) having roughly the same number of variables as imGHUM.

\begin{figure}
    \centering
    \includegraphics[width=0.8\linewidth]{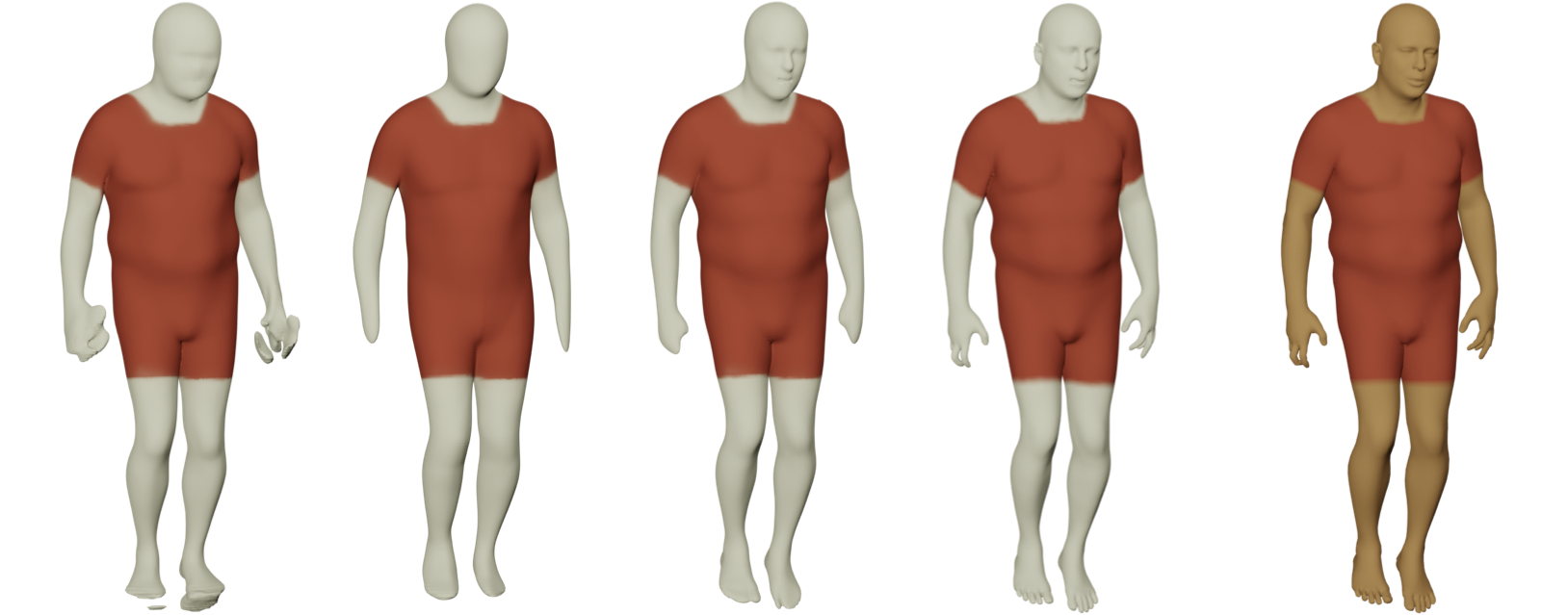}
    \caption{Qualitative comparison with baseline experiments. From left to right: autoencoder, single-part model without and with Fourier input mapping, our multi-part imGHUM, ground-truth GHUM. We use our semantics network to color baseline results. }
    \label{fig:compare_baselines}
    \vspace{2mm}
    \centering
    \includegraphics[width=0.8\linewidth]{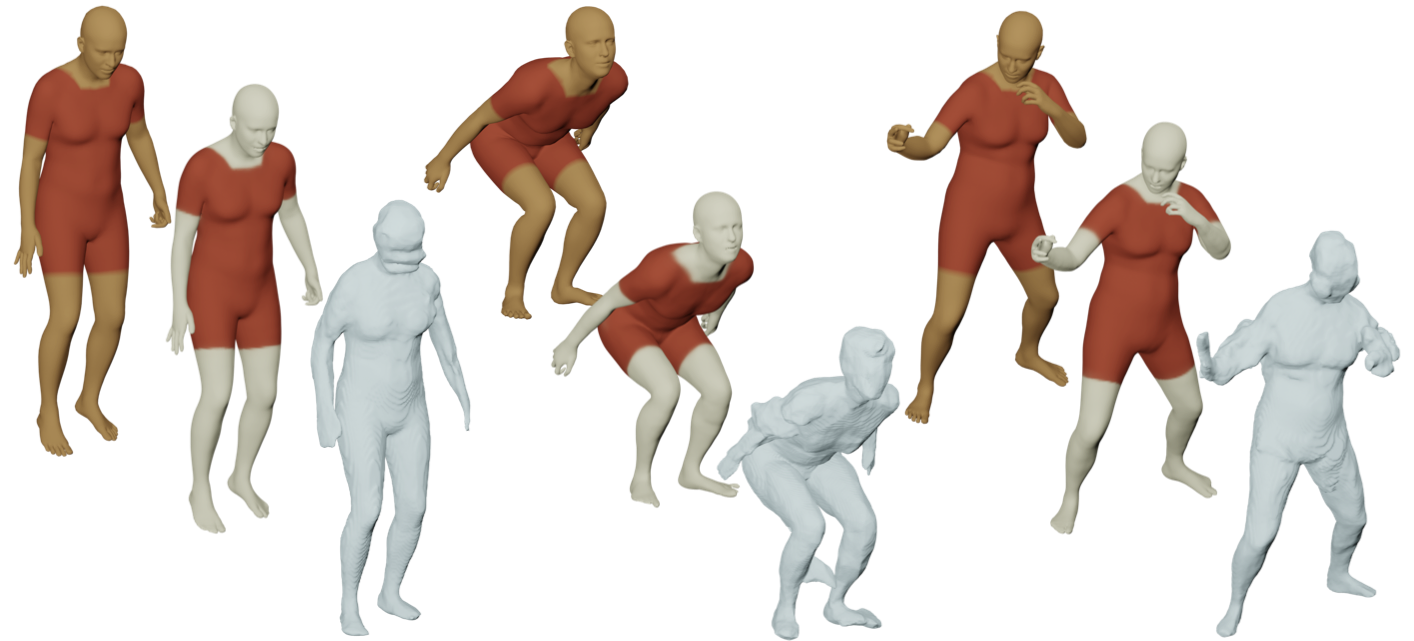}%
    \caption{Comparison with NASA \cite{deng2019nasa} on our single-subject multi-pose dataset. Top to bottom: GT, single-subject imGHUM, and NASA reconstructions. imGHUM better captures global and local geometry, despite using a significantly smaller network version in this experiment. Also numerically our results are superior: IoU ($\uparrow$) $0.962$ (ours) vs.\ $0.839$ (theirs), Ch.\ ($\downarrow$) $0.068\times 10^{-3}$ (ours) vs.\ $3.53\times 10^{-3}$ (theirs), NC ($\uparrow$) $0.985$ (ours) vs.\ $0.903$ (theirs).
    }  
    \label{fig:compare_nasa}
    \vspace{-5mm}
\end{figure}

In tab.~\ref{tab:compare_baselines} we report the metrics for different variants on our test set containing $1\,000$ GHUM samples.
In fig.~\ref{fig:compare_baselines}, we show a side-by-side comparison.
The Fourier input mapping consistently improves results for all variants.
We have also tried higher-dimensional Fourier features but empirically found the basic encoding to work best in our setting.
The auto-encoder produces large artifacts especially in the hand region.
Similar problems, large blobs or missing pieces, can be observed in results from single-part variants, especially for the hands and, less severe, also for the facial region.
These problems, however, are not well captured by globally evaluating the whole shape.
To this end, we evaluate imGHUM and our single-part models specifically for these critical regions, see tab.~\ref{tab:compare_baselines_parts}.
Only imGHUM consistently produces high-quality results also for hands and the face, supporting the proposed architecture choices. 

Next, we compare imGHUM to the recent single-subject multi-pose implicit human occupancy model NASA~\cite{deng2019nasa}. %
With a fixed body shape, we generate $22\,500$ random GHUM full-body training poses and $2\,500$ testing poses from Human3.6M~\cite{Ionescu14pami} and the CMU mocap dataset~\cite{CMUMotionCap2000}, including head and hand poses. Using the original point sampling strategy in NASA, we have trained the network until convergence, based on the original source code.
Please see the supplementary material for details on how we adapted NASA for the GHUM skeleton.
For comparison, we have trained an imGHUM architecture with $2\times$ fewer layers than our full multi-subject model, each with half-dimensionality, using the same dataset.     
Even though GHUM-based NASA has $3\times$ more parameters, our smaller-size single-subject imGHUM still performs significantly better in representing both the global shape and local detail (see hand reconstructions in fig.~\ref{fig:compare_nasa}).
In contrast to NASA, which computes binary occupancy, imGHUM returns more informative
signed distance values which produce smooth decision boundaries and preserve the detailed geometry much better. 
Further key differences to NASA are our considerably simpler architecture that requires far less computation to produce a reconstruction, our semantics, and the carefully chosen learning model (i.e.\ Fourier encoding, second-order losses) that pays particular attention to surface detail.
Moreover, imGHUM additionally models body shape, fingers, and facial expressions using generative latent codes (tab.~\ref{tab:features}).

\begin{figure*}
    \centering
    \includegraphics[width=1\linewidth]{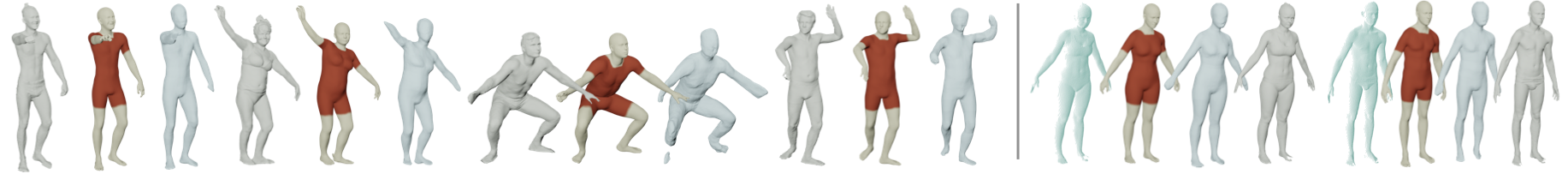}%
    \vspace{-0.5mm}
    \caption{\emph{Left:} Triangle set surface reconstruction (input scan, imGHUM fit, and IF-Net inference from left to right). Numerically, imGHUM fits are better than IF-Net with Chamfer distance ($\downarrow$) $0.156\times 10^{-3}$ (ours) vs.\ $0.844\times 10^{-3}$ (IF-Net), and NC ($\uparrow$) $0.954$ (ours) vs.\ $0.914$ (IF-Net). \emph{Right:} Partial point cloud completion (input point cloud, imGHUM fit, IF-Net, and ground truth scan).}
    \label{fig:surface_reconstruction}
    \vspace{-3mm}
\end{figure*}

\subsection{Applications}
\label{sec:applications}

We apply imGHUM to three key tasks: body surface reconstruction, partial point cloud completion, and dressed and inclusive human reconstruction. \\

\vspace{-3mm}
\noindent{\bf Triangle Set Surface Reconstruction.}
Given a triangle set (`soup') with $n$ vertices $\{\hat\vv\} \in \RR^{3n}$ along with oriented normals $\{\hat\nn\} \in \RR^{3n}$,
we deploy our parametric implicit SDF for surface reconstruction with semantics.
This task is necessary for triangle soups produced by 3D scanners.
To extract the surface from an incomplete scan, we apply a BFGS optimizer to fit $\balpha= (\bbeta_b, \bbeta_f, \btheta)$ such that all vertices $\hat\vv$ are close to the implicit surface $S(\cdot, \balpha) = 0$.
Moreover, we enforce gradients at $\hat\vv$ to be close to  normals $\hat\nn$, and generated off-surface samples to have  distances with the expected signs.
In addition, we sample near surface points with a small distance $\eta$ along surface normals, and enforce $S(\hat\vv \pm \eta\hat\nn, \balpha) = \pm\eta$, as in \cite{i_DeepSDF}. 
Note that all these operations can be easily implemented and are fully differential due to imGHUM being a SDF.
When 3D landmarks are available on the target surface, e.g.\ as triangulated from 2D detected landmarks of raw scanner images, we additionally augment the optimization with landmark losses based on the imGHUM semantics. Please see the supplementary material for details of the losses.

For reference, we also show results on IF-Net~\cite{chibane20ifnet}, a recent method for implicit surface extraction, completion, and voxel super-resolution.
We trained IF-Net with the same pose and shape variation as used for imGHUM -- presumably much more variation than the $2\,183$ scans in the original paper.
In both training and testing we generate $15$K random samples from the observed shape and pass them through IF-Net for surface reconstruction.
Note that IF-Net is using less information compared to our method, but is also solving an easier task as it is not computing a global and semantically meaningful shape code.
An entirely fair comparison is thus not possible.
However, we believe that by comparing with IF-Net, we show that imGHUM is adequate for this task. 
Fig.~\ref{fig:surface_reconstruction} qualitatively shows examples of both imGHUM fits and of IF-Net inference results for $150$ human scans containing $20$ subjects. Our model not only fits well to the volume of the scans but also reconstructs the facial expressions and hand poses. Using landmarks and ICP losses, one could also fit GHUM to the triangle sets. However, our fully  differential imGHUM losses show superior performance over ICP-based GHUM fitting (Chamfer ($\downarrow$) $0.77\times 10^{-3}$, NC ($\uparrow$) $0.921$).
\\

\vspace{-3mm} 
\noindent{\bf Partial Point Cloud Completion.}
Another relevant task for many applications is shape completion.
Here we show surface reconstruction and completion from partial point clouds as recorded e.g.\ using a depth sensor.
We synthesize depth maps from A-posed scans of $10$ subjects from the Faust dataset~\cite{bogo2014faust} using the intrinsics and the resolution of a Kinect V2 sensor.
To complete the partial view, we search for the $\balpha$ such that all points from the depth point cloud are close to imGHUM's zero-level-set.
We sample additional points along surface normals (estimated from depth image gradients) and enforce estimated distances by imGHUM to be close to true distances.
We also sample points in front of the depth cloud and around it and enforce their $L_{l}$ label loss.
Finally, we also supervise the estimated normals.
We do not rely on landmarks or other semantics in this experiment.

We show IF-Net~\cite{chibane20ifnet} results for comparison.
We trained IF-Net specifically for this task while we use the same imGHUM for all experiments. Our reconstructions are numerically better with Chamfer distance ($\downarrow$) 
$0.103\times10^{-3}$ (ours) vs.\ $0.315\times10^{-3}$ (theirs) and NC ($\uparrow$) $ 0.962$ (ours) vs.\ $0.936$ (theirs).
Qualitatively, our results contain much more of the desirable reconstruction detail, especially for hands and faces, see fig.~\ref{fig:surface_reconstruction}, right. %
Note, again, that IF-Net only reconstructs a surface while we recover the parametrization of a body model, a considerably harder task.\\

\begin{figure}
    \centering
    \includegraphics[width=1\linewidth]{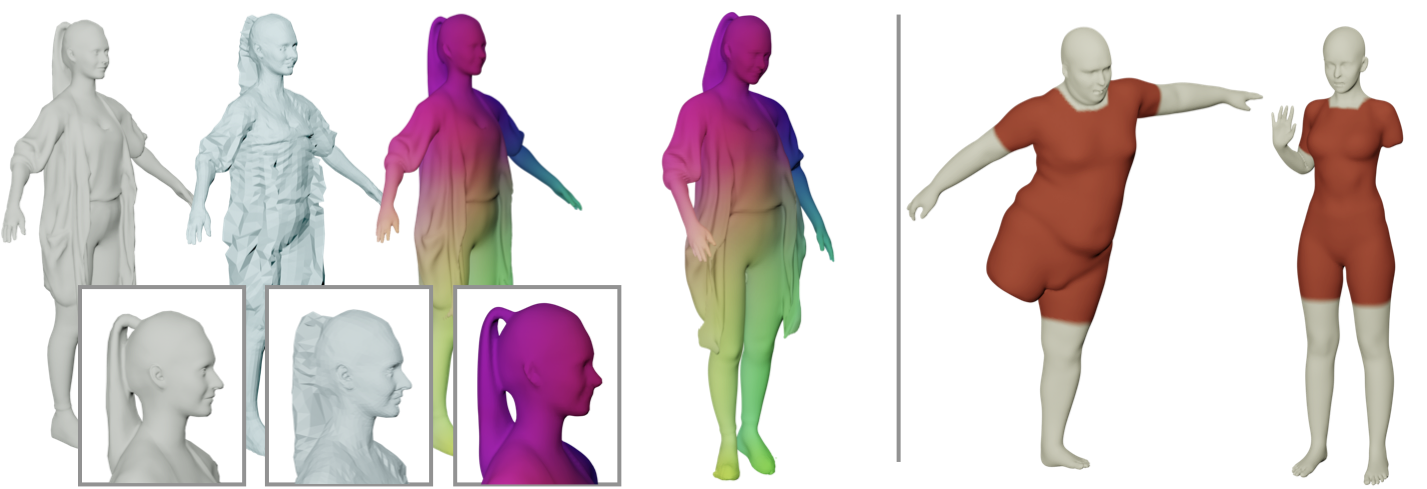}%
    \caption{From left to right: scan, GHUM template mesh ACAP registration, imGHUM+residual fit (color-scale represents semantics), reposed imGHUM+residual, imGHUM+residual fits to people with limb differences. 
    In contrast to the fitted template mesh, imGHUM+residual successfully models topologies different from the plain human body and captures more geometric detail.}
    \label{fig:clothed_recon}
    \vspace{-1mm}
\end{figure}

\vspace{-4mm}
\noindent{\bf Dressed and Inclusive Human Modeling.}
imGHUM is template-free which is a valuable property for future developments.
While this work deals primarily with the methodology of learning a generative implicit human model -- in itself a complex and novel task -- we also give an outlook for possible future directions.
Building a detailed model of the human body shape including hair and clothing, or learning inclusive models could be such directions.
However, currently the data needed for building such models does not exist at large enough scale.
To demonstrate that imGHUM is a valuable building block for such models, we leverage it as an inner layer for personalized human models.
Concretely, we augment imGHUM with a light-weight residual SDF network, conditioned on the output of imGHUM, both the signed distances and semantics.
We estimate the residual model using the same learning scheme as for imGHUM, but limit training to a single scan.
The final output models the human with layers, including the inner body shape represented with imGHUM and the personalization (hair, clothing, non-standard body topology) as residuals, \cf fig.~\ref{fig:clothed_recon}.
This layered representation can be reposed by changing the parameterization of the underlying imGHUM.
Hereby, the residual model acts as a fitted layer around imGHUM and deforms according to the distance and semantic field defined by imGHUM. Please see the supplementary material for more examples, a numerical evaluation, and implementation details.

%% file: sections/conclusion.tex
We introduced imGHUM, the first 3D human body model, with controllable pose and shape, represented as an implicit signed distance function.
imGHUM has comparable representation power to state-of-the-art mesh-based models and can represent significant variations in body pose, shape, and facial expressions, as well as underlying, precise, semantics.
imGHUM has additional valuable properties, since its underlying implicit SDF represents not only the surface of the body but also its neighborhood, which e.g.\ enables collision tests with other objects or efficient distance losses. 
imGHUM can be used to build diverse, fair models of humans who may not match a standard template. 
This paves the way for transformative research and inclusive applications like modeling clothing, enabling immersive virtual apparel try-on, or free-viewpoint photorealistic visualization. Our models are available for research~\cite{imghumcode}.

%% file: sections/suppl.tex
\appendix 

In this supplementary material, we detail our implementation and used architectures, show further results, discuss further ablations, and give more details on our experiments and comparisons.
We also demonstrate how imGHUM can be used for differentiable rendering.

\section{Implementation Details}
In the following, we detail the implementation of imGHUM.
We specify the used hyper-parameters and the architectures used in the ablation experiments.
Finally, we give running times for imGHUM mesh extraction via Marching Cubes~\cite{lewiner2003efficient}.

\vspace{-2mm}
\paragraph{Hyper-parameters.}
We train imGHUM with a batch-size of $32$, each of which contains $32$ instances of $\balpha$ paired with $512$ on surface, $256$ near surface, and $256$ uniform samples for each instance.
Our loss is composed as
\begin{gather}
    L = \lambda_{o_1} L_{o_1} + \lambda_{o_2} L_{o_2} + \lambda_{e}L_{e} +\lambda_{l} L_{l} +\lambda_{s} L_{s},
\end{gather}
where $L_{o_1}$ refers to the first part of $L_{o}$ (distance) and $L_{o_2}$ to the second part (gradient direction), respectively, and $L_{s}$ refers to the semantics loss.
We choose $\lambda_{o_1} = 1$, $\lambda_{o_2} = 1$, $\lambda_{e} = 0.1$, $\lambda_{l} = 0.1$, and $\lambda_{s} = 0.5$.
Empirically we found that linearly increasing $\lambda_{o_1}$ to $50$ over 100K iterations leads to perceptually better results. 
We train imGHUM until convergence using the Adam optimizer \cite{kingma2014adam} with a learning rate of $0.2\times10^{-3}$ exponentially decaying by a factor of $0.9$ over 100K iterations.

\vspace{-2mm}
\paragraph{Architectures.}

The following architectures have been used for the baseline experiments:
The single-part network has been used as described in the main paper totaling in $2.01$M parameters.
The deeper single-part network uses $10$ instead of $8$ layers, resulting in $2.53$M parameters.
The autoencoder is composed from a PointNet++ \cite{qi2017pointnet++} encoder and our single-part decoder with a total number of parameters of $3.91$M.
The encoder consists of three PointNet++ set abstraction modules and two 512-dimensional fully-connected layers with ReLU activation.

\begin{figure*}
    \centering
    \includegraphics[width=0.78\linewidth]{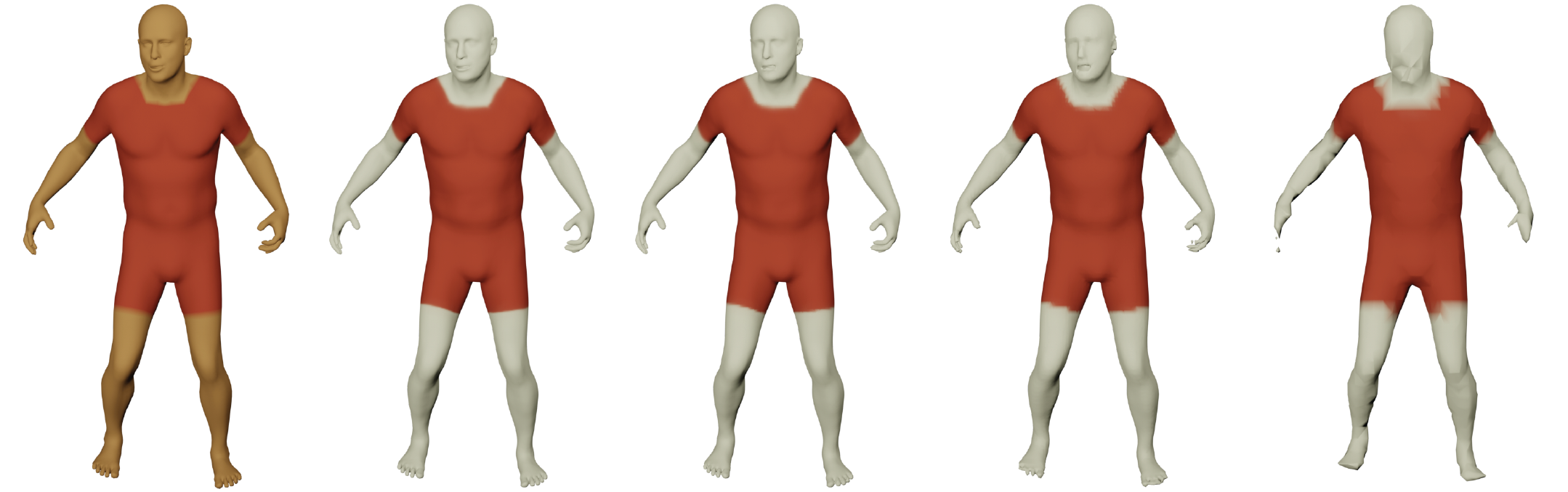}
    \caption{imGHUM mesh reconstructions in different resolutions. Left to right: ground-truth shape, $512^3$,  $256^3$, $128^3$, $64^3$.}
    \label{fig:mc}
    \vspace{-1mm}
\end{figure*}

\vspace{-2mm}
\paragraph{Running Times.}
We extract meshes from imGHUM using using Octree sampling.
Reconstructing a mesh in its bounding box and with a maximum grid resolution of $256^3$ takes on average $1.08$s using a NVIDIA Tesla V100.
Hereby, the network query time sums up to $0.44$s and Marching Cubes~\cite{lewiner2003efficient} (on CPU) takes $0.34$s.
The rest of the time is used by identifying the bounding box through probing ($0.17$s), Octree logic ($0.05$s), and transforming the samples to the part reference frames ($0.07$s). 
We query imGHUM in batches with a maximum batch-size of $64^3$ samples, where one full batch takes on average $0.13$s to compute.
The resulting meshes feature approximately $100$K vertices and $200$K facets.
Note that imGHUM allows creating meshes in arbitrary resolutions and can be queried and also rendered (\cf \S\ref{sec:diff_render}) without generating an explicit mesh.
For reference, we show imGHUM mesh reconstructions in different resolutions in fig.~\ref{fig:mc}.

\section{Results}
\begin{figure*}
    \centering
    \includegraphics[width=1\linewidth]{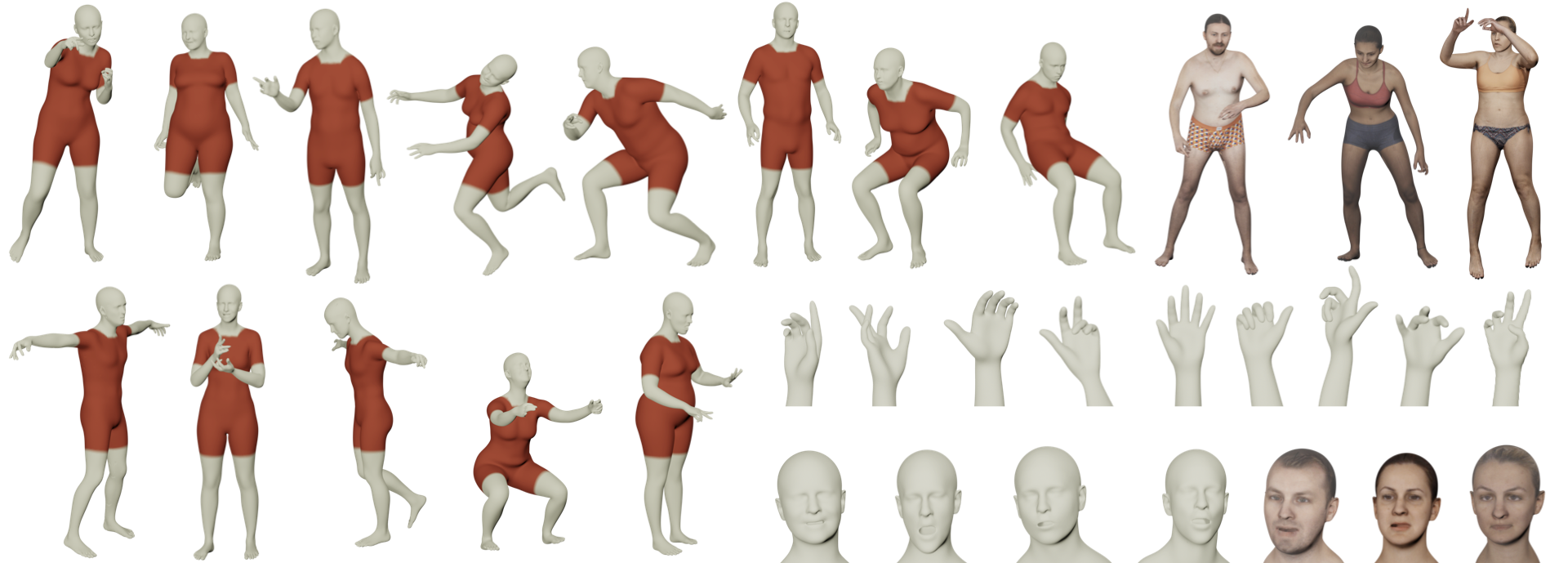}
    \caption{Random imGHUM full-body and part instances sampled from imGHUM's generative latent codes. On the right, we show textured examples. Texturing and binary coloring is enabled by imGHUM's semantics.}
    \label{fig:generative_suppl}
    \vspace{5mm}

    \centering
    \includegraphics[width=1\linewidth]{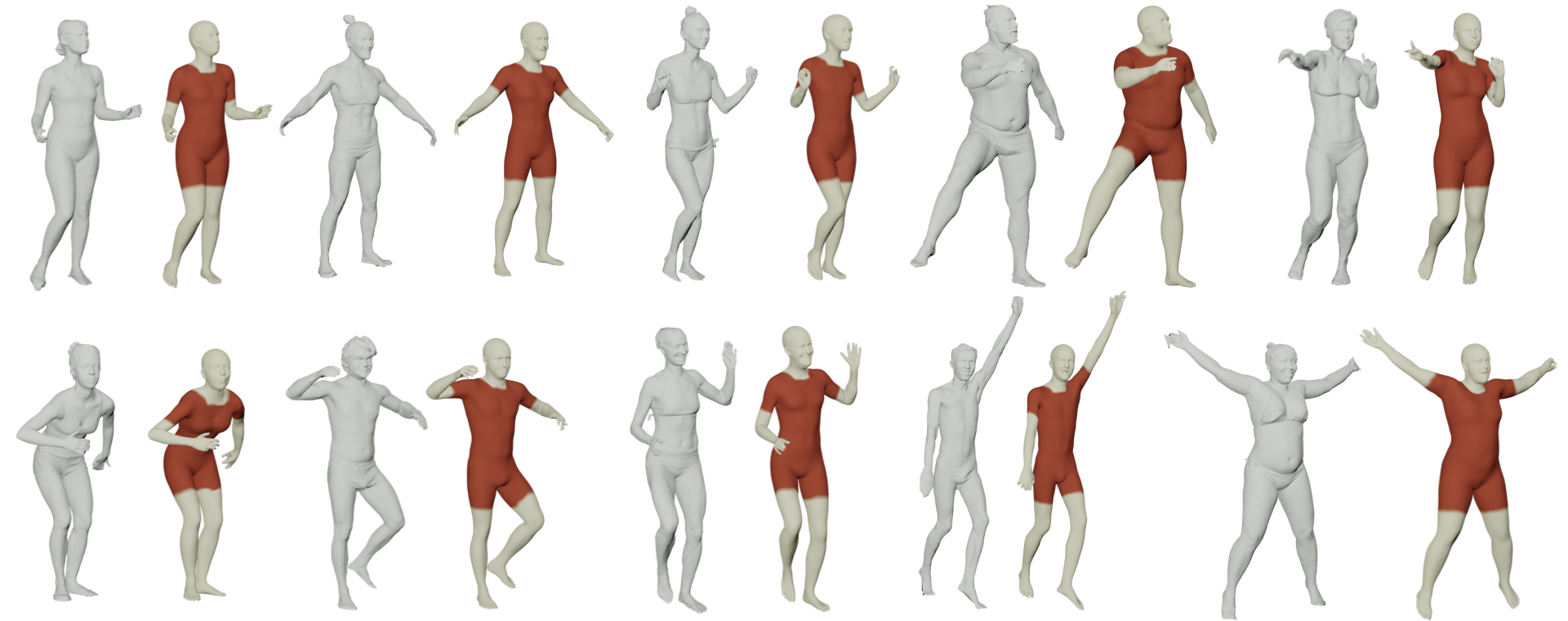}
    \caption{More examples for the triangle set surface reconstruction experiment. Each pair shows the ground truth scan (left) and our reconstruction (right). Notice the reconstructed facial expressions and hand poses.}
    \label{fig:fitting_remaining}
    \vspace{5mm}

    \centering
    \includegraphics[width=1\linewidth]{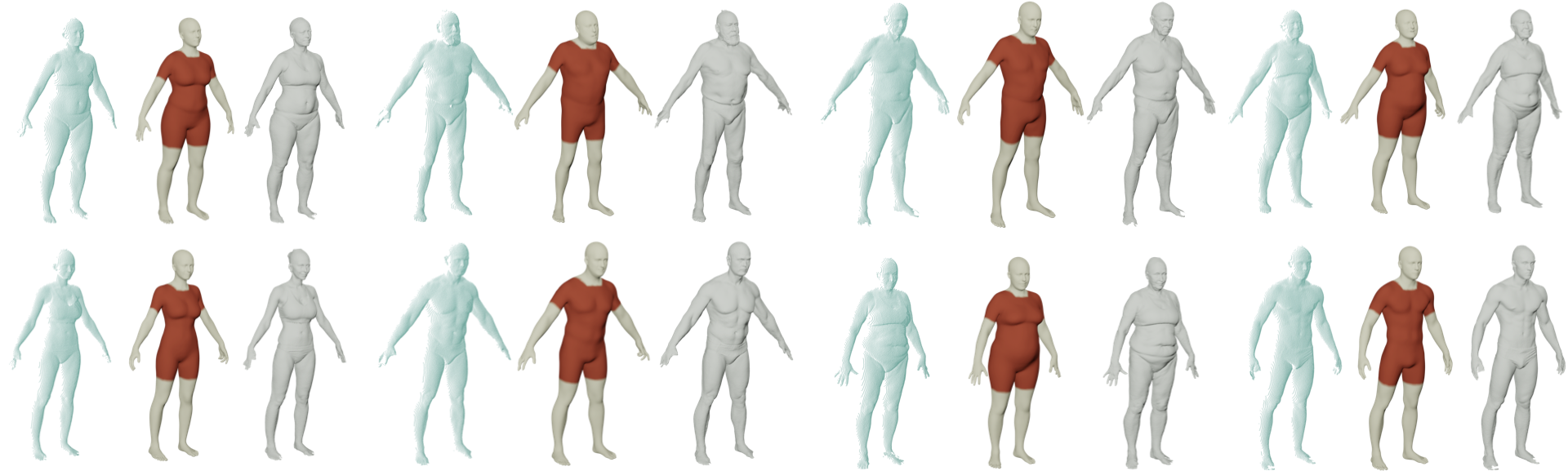}
    \caption{Remaining partial point cloud completion results. Left to right: input point cloud, imGHUM fit, and ground truth scan.}
    \label{fig:completion_remaining}
\end{figure*}

In this supplemental material we show additional results for our application experiments (fig.~\ref{fig:fitting_remaining}, \ref{fig:completion_remaining}, \ref{fig:residual_dressed}, \ref{fig:residual_disabled}). Additionally, fig.~\ref{fig:generative_suppl} displays a large number of imGHUM instances with great variety in poses, shapes, hand poses, and facial expressions sampled from imGHUM's generative latent space.
This demonstrates once more that imGHUM's level of detail, expressiveness and generative power is on par with state-of-the-art mesh-based models.
Moreover, imGHUM can additionally be queried at arbitrary resolutions and spatial locations and models not only the surface, but also the space around the person.

\section{Ablations}
In this section, we report further results of our dataset ablation experiment and results of an additional ablation study on joint rotation parameterization.

\vspace{-2mm}
\paragraph{Dataset.}

\begin{table}
    \begin{center}
    \footnotesize
    \begin{tabular}{l|ccc}  
    Model & IoU $\uparrow$ & Chamfer $\times 10^{-3}$ $\downarrow$ & NC $\uparrow$\\
    \hline
    \hline
    Only scan registrations & 0.901 & 0.091 & 0.975 \\ 
    \textbf{imGHUM} & \textbf{0.932} & \textbf{0.040}  & \textbf{0.984} \\ 
    \end{tabular}
    \end{center}
    \caption{Numerical comparison of imGHUM trained with different data distributions evaluated on the registration test-set.}
    \label{tab:registrations}
\end{table}

\begin{table}
     \begin{center}
     \footnotesize
     \begin{tabular}{l|ccc}  

     Model & IoU $\uparrow$ & Chamfer $\times 10^{-3}$ $\downarrow$ & NC $\uparrow$\\
     \hline
     \hline
     Only scan registrations  & 0.834  & 2.561 & 0.942 \\ 
     \textbf{imGHUM}    & \textbf{0.969} & \textbf{0.036} & \textbf{0.989} \\
     \end{tabular}
     \end{center}
     \caption{Numerical comparison of imGHUM trained with different data distributions evaluated on the GHUM samples test-set.}
     \label{tab:ghumsampling}
     \vspace{-1mm}
\end{table}

In the main paper we have shown that imGHUM benefits from being trained on both samples of GHUM and additionally on As-Conformal-As-Possible (ACAP) registrations of a corpus of human scans. 
While training only on scan data can represent the distribution of the scans well (tab.~\ref{tab:registrations}), it does not generalize sufficiently to poses that are not covered in this limited training set, as we show in tab.~\ref{tab:ghumsampling}.

In fig.~\ref{fig:regnoreg}, we qualitatively show the effect of fine-tuning with scan data. Please note the increased level of detail in the faces and the enhanced soft-tissue deformation.

\vspace{-2mm}
\paragraph{Rotation Representations.}

In tab.~\ref{tab:rotations}, we report metrics for imGHUM using different rotation representations for joint rotations $\btheta$.
We have experimented with Euler angles, basic $\sin$, $\cos$ Fourier mapping~\cite{Zanfir2020Weakly}, and the recently proposed 6D representation \cite{zhou2019continuity}.
Perhaps surprisingly, we found only minor differences in imGHUM's representational power using different rotation representations, both qualitatively and quantitatively.
We, therefore, use Euler angles in this work as it is the most compact representation.

\begin{table}
    \begin{center}
    \footnotesize
    \begin{tabular}{l|ccc}  

    Model & IoU $\uparrow$ & Chamfer $\times 10^{-3}$ $\downarrow$ & NC $\uparrow$\\
    \hline
    \hline
    6D & \textbf{0.969} & 0.044 & \textbf{0.989} \\ 
    $\sin$, $\cos$ & 0.967 & 0.046 & 0.988 \\ 
    Euler & \textbf{0.969} & \textbf{0.036} & \textbf{0.989}\\ 
    \end{tabular}
    \end{center}
    \caption{Numerical comparison of imGHUM models using different representations for joint angles evaluated on the GHUM samples test-set.}
    \label{tab:rotations}
\end{table}

\section{Applications}
In the following, we explain the losses used in our triangle set surface reconstruction experiment, detail the residual model of the dressed and inclusive modeling experiment, and finally introduce another application namely pose estimation from silhouettes using differentiable rendering.

\subsection{Triangle Set Surface Reconstruction}
We describe our triangle set surface reconstruction experiment in the main paper (\S3.3) and show more examples here in fig.~\ref{fig:fitting_remaining}. Our imGHUM reconstructions are performed under a weighted combination of losses as 
\begin{gather}
    \small
    \label{eq:imghum_fitting_losses}
    \min_{\balpha} L_o(\balpha) + L_l^+(\balpha)  + L_l^-(\balpha) \\
    \small
    L_o(\balpha) = \frac{1}{n}\sum_i |S(\hat\vv_i, \balpha)| + \| \nabla_{\hat\vv_i}S(\hat\vv_i, \balpha) - \hat\nn_i \|  \\
    \small
    L_l^+(\balpha) = \frac{1}{n}\sum_{i}\bigl(\phi(k S(\hat\vv_i + \gamma_i\hat\nn_i, \balpha)) - 1\bigr)^2 \\
    \small
    L_l^-(\balpha) = \frac{1}{n}\sum_{i}\bigl(\phi(k S(\hat\vv_i - \gamma_i\hat\nn_i, \balpha))\bigr)^2,
\end{gather}
where $L_o$ is a surface sample loss (similar to eq.~2 in the main paper), and $L_l^+, L_l^-$ are sign classification losses
defined for points sampled along and opposite to the normals respectively ($\gamma_i \in [0, 0.05]$ is a Gaussian sampled distance).

Enabled by the implicit semantics of imGHUM, we can additionally exploit landmark losses as, 
\begin{gather}
    \small
    L_j(\balpha) = \frac{1}{|M_j|}\sum_{i \in M_j}\bigr\|\TT_i(\balpha) \centers_i(\balpha) - \mm_{j,i} \bigl\|^2 \\
    \small
    L_s(\balpha) = \frac{1}{|M_s|}\sum_{i \in M_s}\bigr\|\CC(\mm_{s,i}, \balpha) - \bar{\mm}_{s,i} \bigr\|^2,
\end{gather}
where $M_j = \{\mm_{j}\}$, $M_s = \{\mm_{s}\}$ are a collection of 3D landmarks defined over the joints and the surface,  respectively.
$\bar{\mm}_{s}$ are the corresponding surface landmarks defined on the canonical mesh $\XX(\balpha_0).$
$L_j$ aligns the transformed joint centers with the joint landmarks. 
The surface landmarks loss $L_s$ queries the semantics for the ground-truth surface landmarks $\mm_{s}$ conditioned on $\balpha$.
The semantics describe the position of the landmarks $\mm_{s}$ w.r.t.\ the canonical mesh, and thus should match their correspondences $\bar{\mm}_{s}$.

Given a triangle set mesh, one could also fit GHUM with landmarks and ICP losses. However, we note that imGHUM is not only able to perform equivalently on the landmark losses to mesh-based representations, but also exploits more information of the triangle set with its differential losses (eq.~\eqref{eq:imghum_fitting_losses}) compared to ICP. The process of finding the nearest point for ICP at each optimization iteration is non-differentiable and the accuracy of the nearest point correspondences are highly sensitive to the initialization. In contrast, our imGHUM losses are fully differential everywhere and also exploit additional information encoded in the surface normals and the sign labels. Numerical comparisons are reported in \S3.3 of the main paper.

\subsection{Dressed and Inclusive Human Modeling}
\begin{figure*}
    \centering
    \includegraphics[width=1\linewidth]{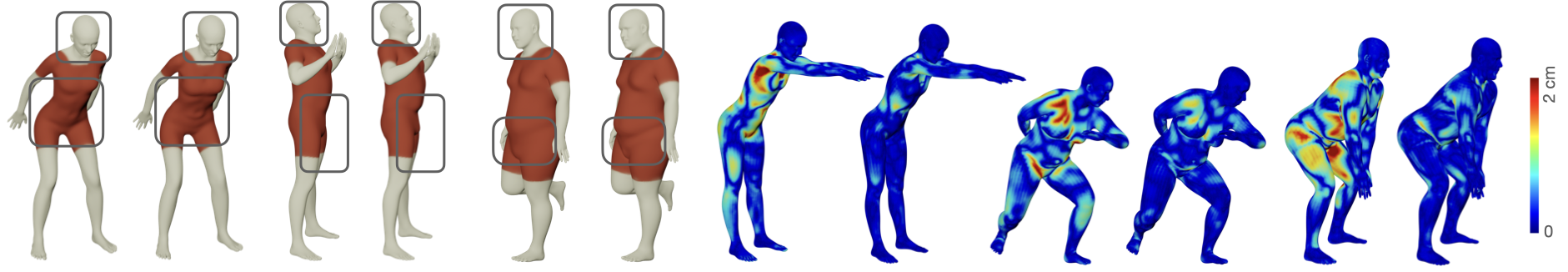}
    \caption{Added detail after fine-tuning on the registration dataset.  We show imGHUM reconstructions before fine-tuning (left) and after fine-tuning (right) qualitatively and using error heat-maps (red means $\ge2$cm). Please pay attention to the faces, body shapes, and soft-tissue deformations (digital zoom in recommended).}
    \label{fig:regnoreg}
    \vspace{4mm}
    \centering
    \includegraphics[width=1\linewidth]{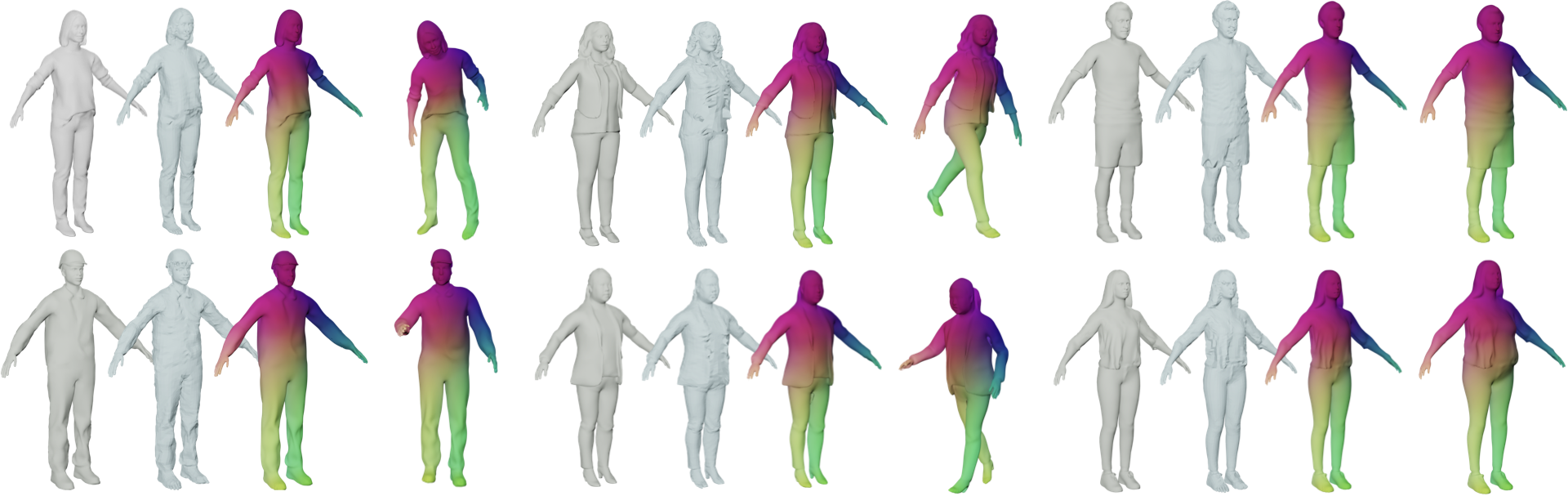}
    \caption{Dressed human modeling. From left to right: Scan, GHUM ACAP mesh registration, imGHUM+residual fit, reposed or reshaped imGHUM+residual. imGHUM+residual accurately explains all detail present in the input scan. GHUM ACAP mesh registrations have difficulties with complicated and layered structures. By changing the parameterization of the underlying imGHUM, we can repose and reshape the personalized models. The color-scale represents imGHUM semantics and thus correspondences between different instances.}
    \label{fig:residual_dressed}
    \vspace{4mm}
    \centering
    \includegraphics[width=1\linewidth]{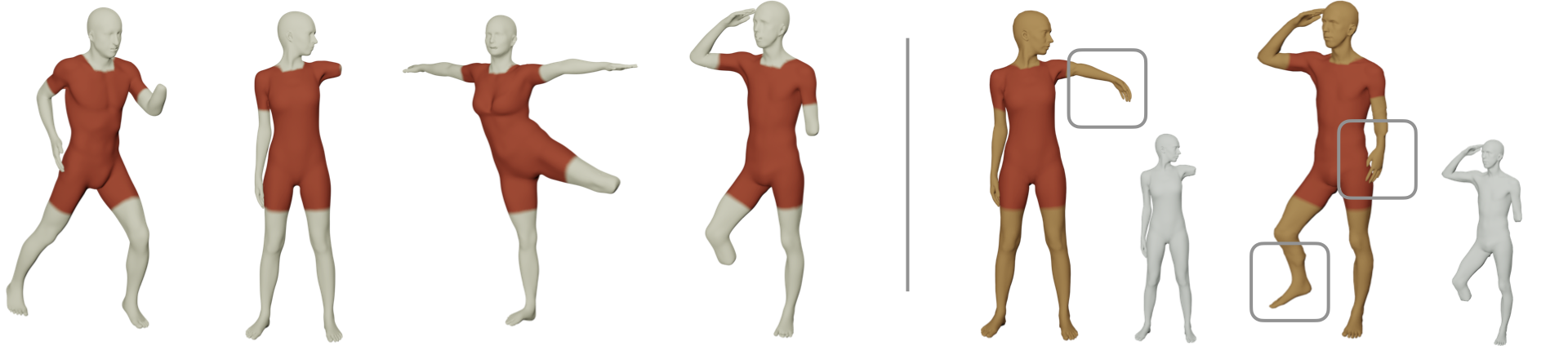}
    \caption{Inclusive human modeling. \emph{Left:} imGHUM+residual can explain body shapes that do not match the standard template. \emph{Right:} GHUM ACAP mesh registrations fail to explain these body shapes. For reference, we show ground truth scans in small. Missing limbs are deformed but still present.}
    \label{fig:residual_disabled}
    \vspace{4mm}
    \centering
    \includegraphics[width=1\linewidth]{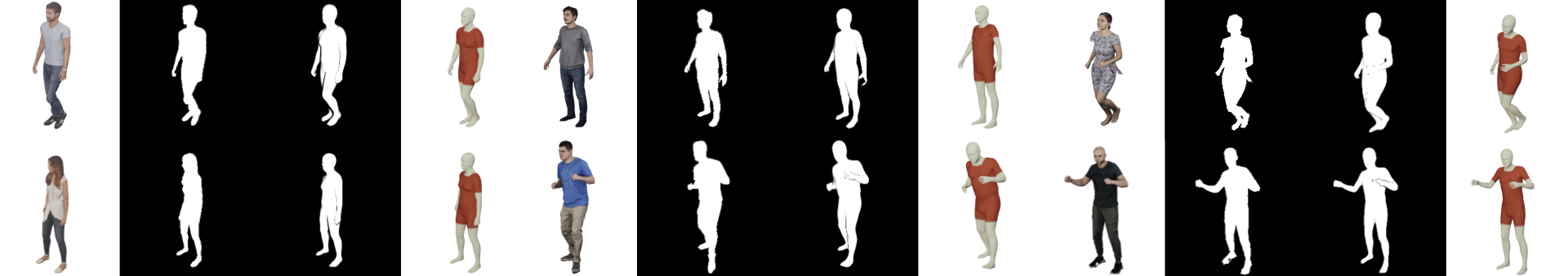}
    \caption{Visual 3D reconstruction of imGHUM using differentiable 
    silhouette and landmark losses. Left to right: image, observed silhouette, estimated silhouette, imGHUM reconstruction. By using a silhouette loss, we are able to accurately reconstruct body shapes.}
    \label{fig:image_fitting}
\end{figure*}

In the following, we detail our dressed and inclusive modeling experiment from the main paper.
We also show more results in fig.~\ref{fig:residual_dressed}.
In order to learn a personalized shape of a given scan, we augment imGHUM with an MLP $\hat S$ consisting of four 256-dimensional layers.
Each layer is followed by Swish nonlinear activation, and a skip connection is added to the middle layer.
$\hat S$ modulates the signed distance field of the body to match the scan. 
These distance residuals could come from clothing, hair, other apparel items, or any divergence from the standard human template.
We condition the output signed distance of the scan with both the distance and semantics fields of the body defined by imGHUM:
\begin{equation}
    \hat s = \hat S(S(\pp, \balpha)) = \hat S(s, \cc).
\end{equation}
We first fit imGHUM to the scan, similar to the trinagle set surface reconstruction experiment. Next, we train $\hat S$ on top of it. The training process is similar to imGHUM with the difference that we sample points from a single scan containing the desired personalizations. 
We only train the residual while keeping imGHUM fixed. We, therefore, have both the underlying human body and the personalized shape modeled separately as layers. 
We train a separate instance of $\hat S$ for each scan observation.
Learning a combined model using an auto-decoder style learning scheme is possible but beyond the scope of this work.

We show two categories of personalizations: dressed humans and humans with limb differences.
We compare imGHUM+residual with mesh-based GHUM ACAP registrations.
In contrast to our template-free imGHUM+residual model, GHUM ACAP registrations have difficulties in explaining complex and layered structure and unsurprisingly fail entirely for large structural changes.
We fit to scans of ten subjects with limb differences and 30 dressed human scans.
Numerically, imGHUM+residual performs better than GHUM ACAP registrations with Chamfer distance $0.014\times10^{-3}$ (ours, limb differences) / $0.018\times10^{-3}$ (ours, dressed) versus $1.393\times10^{-3}$ (GHUM ACAP, limb differences) / $0.021\times10^{-3}$ (GHUM ACAP, dressed) and Normal Consistency $0.993$ (ours, limb differences) / $0.990$ (ours, dressed) versus $0.984$ (GHUM ACAP, limb differences) / $0.976$ (GHUM ACAP, dressed).
imGHUM+residual is especially superior in explaining the scans of people with limb differences, due to large structural differences compared to the GHUM template mesh.
Also qualitatively imGHUM+residual explains much more of the detail present in the input scans, see fig.~\ref{fig:residual_dressed} and fig.~\ref{fig:residual_disabled}.

\subsection{Differentiable Rendering}
\label{sec:diff_render}
A benefit of imGHUM's SDF representation is the potential for rendering using sphere tracing~\cite{hart1996sphere}. During ray tracing the surface is located by stepping from the camera along a ray until a surface is passed.
In sphere tracing the save step length is given by the current minimal distance to any point on the surface, i.e. the SDF value at the current location.
For inexact SDFs, one can take a damped step to reduce the likelihood of over-shooting. Using this technique we can render among other things: imGHUM depth maps, normal maps, and semantics.
Hereby, each pixel contains the last queried value of its corresponding camera ray.
In the following, we compute differentiable binary silhouettes via sphere tracing and fit imGHUM to images using a silhouette alignment loss.

We implement differentiable approximate sphere tracing by taking a fixed number of steps.
Concretely, we step $T=15$ save steps into the SDF in the direction of each camera ray.
At each final point $\pp_T$ of each camera ray, we query the signed distance value and generate the binarized pixel as:
\begin{equation}
    \small
    b = \frac{1}{\eta S(\pp_T, \balpha)^2 + 1},
\end{equation}
with $\eta=5000$ in our experiment. $b$ is differentiable w.r.t.\ $\balpha$ and thus can be used in optimization losses. 
We formulate a standard silhouette overlap loss and a sparse 2D joint landmark loss and use both to fit imGHUM to image evidence.
Fig.~\ref{fig:image_fitting} shows results of fitting imGHUM to image silhouettes.

\section{Details on Compared Methods}

As reported in the main paper, we change NASA~\cite{deng2019nasa} in contrast to their original version.
Firstly, we train NASA based on the GHUM skeleton containing $63$ parts.
Originally, NASA was trained on SMPL containing only $24$ parts.
Another difference is the topology of GHUM.
In contrast to SMPL, GHUM features an oral cavity that is also represented in our training data.
Summarizing, we deploy NASA for a higher-dimensional model and thus a harder task.
For a fair comparison, we therefore use a larger and deeper architecture with eight $64$-dimensional fully-connected layers for each part instead of the original four $40$-dimensional layers.
The new architecture features $1.92$M parameters (original version has $0.38$M) and has shown significantly better representation power.
In contrast, we use a much smaller imGHUM architecture in this experiment.
imGHUM has been originally designed to also explain shape variation and facial expressions.
Since this experiment only features variation in pose, we can use a much smaller version.
We use $2\times$ fewer layers in each part, each with half-dimensionality, resulting in only $0.64$M parameters.
This smaller-size imGHUM still performs significantly better than NASA in our experiments.

We have trained IF-Net~\cite{chibane20ifnet} based on their original source code.
Specifically, we use IF-Net for point clouds with $128^3$ resolution featuring $2.6$M parameters.
We also follow their sampling and resizing strategy, such that the input point cloud always has a maximum side length of one unit.
Finally, we train IF-Net task-specific (for full and partial point clouds), while we use the same imGHUM in all our comparisons.